\documentclass[a4paper,fleqn]{cas-dc}

\usepackage[authoryear,longnamesfirst]{natbib}
\usepackage{url}

\usepackage{graphicx}
\graphicspath{{fig/}}
\DeclareGraphicsExtensions{.pdf,.jpeg,.png}
\usepackage{subfig}

\def\tsc#1{\csdef{#1}{\textsc{\lowercase{#1}}\xspace}}
\tsc{WGM}
\tsc{QE}
\begin{document}
\let\WriteBookmarks\relax
\def\floatpagepagefraction{1}
\def\textpagefraction{.001}

\shorttitle{BezierSeg}    

\shortauthors{Haichou~Chen et~al.}  

\title [mode = title]{BezierSeg: Parametric Shape Representation for Fast Object Segmentation in Medical Images}  

%\tnotemark[<tnote number>] 

%\tnotetext[<tnote number>]{<tnote text>} 


\author[1]{Haichou~Chen}[style=chinese,orcid=0000-0003-2639-8478]

\fnmark[1]

\ead{movic.chen@gmail.com}

%\ead[url]{<URL>}

%\credit{<Credit authorship details>}

\author[1]{Yishu~Deng}[style=chinese]
\fnmark[1]
\ead{dengys@sysucc.org.cn}

\author[1]{Bin~Li}[style=chinese]
\ead{libin@sysucc.org.cn}
\author[1]{Zeqin~Li}[style=chinese]
\ead{zeqin.le@gmail.com}
\author[1]{Haohua~Chen}[style=chinese]
\ead{fduivchh@gmail.com}

\author[1]{Bingzhong~Jing}[style=chinese]

\cormark[1]
\ead{jingbzh@sysucc.org.cn}

\author[1]{Chaofeng~Li}[style=chinese]

\cormark[1]
\ead{lichaofeng@sysucc.org.cn}

\affiliation[1]{organization={Sun Yat-sen University Cancer Center, Collaborative Innovation Center for Cancer Medicine, State Key Laboratory of Oncology in South China and Guangdong Key Laboratory of Nasopharyngeal Carcinoma Diagnosis and Therapy},
            addressline={651 Dongfeng Road East}, 
            city={Guangzhou},
%          citysep={}, % Uncomment if no comma needed between city and postcode
            postcode={510060}, 
            state={Guangdong},
            country={China}}

%\author[<aff no>]{<author name>}[<options>]

% Footnote of the second author

% Email id of the second author
%\ead{}



%\affiliation[<aff no>]{organization={},
%%          citysep={}, % Uncomment if no comma needed between city and postcode

\cortext[1]{Corresponding author}

\fntext[1]{Co-first author}


\begin{abstract}
Delineating the lesion area is an important task in image-based diagnosis. Pixel-wise classification is a popular approach to segmenting the region of interest. However, at fuzzy boundaries such methods usually result in glitches, discontinuity, or disconnection, inconsistent with the fact that lesions are solid and smooth. To overcome these undesirable artifacts, we propose the \emph{BezierSeg} model which outputs bezier curves encompassing the region of interest. Directly modelling the contour with analytic equations ensures that the segmentation is connected, continuous, and the boundary is smooth. In addition, it offers sub-pixel accuracy. Without loss of accuracy, the bezier contour can be resampled and overlaid with images of any resolution. Moreover, a doctor can conveniently adjust the curve's control points to refine the result. Our experiments show that the proposed method runs in real time and achieves accuracy competitive with pixel-wise segmentation models.
\end{abstract}


% Keywords
\begin{keywords}
bezier curves \sep  deep learning \sep object segmentation \sep
\end{keywords}

\maketitle

% Main text
%\section{}\label{}

\section{Introduction}
Image segmentation is a fundamental task in medical image processing. According to the statistics of the Grand Challenges~\citep{grand_challenge} competition in biomedical image analysis, there are 10, 14, 13 tasks related to image segmentation in 2018, 2019 and 2020 respectively. Deep learning has achieved a prodigious success in image processing with unprecedented accuracy and generality~\citep{Wang_2019,ronneberger2015u,chen2018mixnet,heker2020joint}. As a result, nowadays most newly developed biomedical annotation tasks employ deep learning segmentation models such as Unet++~\citep{zhou2018unet++}, DeepLab v3+~\citep{chen2018encoder}, etc. However, these pixel-based deep segmentation models may not fully satisfy the need in medical application due to its own specialty: firstly, most targets are solid objects with continuous and confined boundaries, for instance, skin lesions. Secondly, biomedical annotation task is normally not the final task but a intermediate step of the whole therapy process. Further manipulations might be applied to the annotations for down-steam tasks like diagnose, lesion measurement, radiotherapy, etc. As for pixel-based algorithm, the output may not have a clear contour, or have burrs along the contour. Of course, one can do post-processing, like binarizing the segmentation heat map and pass it to \textit{OpenCV}~\citep{opencv_library} to find the contour. However, the output might contain multiple contours or a too complex contour for a single object. The first case needs additional post-processing like Non-Maximum Suppression (NMS) to get a most-possible contour, and the second case needs contour-simplify algorithm to get a reasonable contour for doctors to further process. These drawbacks show that the pixel-based segmentation models might not be a straightforward solution for biomedical annotations. Moreover, all pixel-based segmentation models need upsampling operations that further slowdown the whole process.
In this paper, we propose a contour-based model - \emph{BezierSeg}, an end-to-end segmentation model that does not need upsample operations and can output a clear bezier contour directly. Bezier curve has been widely used in many enterprise design software because of its user-friendly properties. Therefore, the predicted bezier contour of the lesion area can be easily manipulated by doctors for further study.
\section{Related Works}
In deep learning, segmentation usually includes semantic segmentation and instance segmentation. The aim of semantic segmentation is to give pixel-wise classification results for the whole input image, and instance segmentation outputs bounding boxes for the detected objects and pixel-wise segmentation within the bounding boxes. 

\subsection{Pixel-based Segmentation}
Most pixel-based semantic segmentation models are fully convolutional networks (FCN), such as U-Net~\citep{ronneberger2015u}, PSPNET~\citep{zhao2017pyramid}, BiSeNet~\citep{yu2018bisenet} and DeepLab v3+~\citep{chen2018encoder}. U-Net has been widely used in biomedical segmentation problems. DeepLab v3+ is a cutting-edge semantic segmentation model developed by Google, which employs the atrous spatial pyramid pooling (ASPP) layer to exploit multi-scale features.
Two stage methods perform instance segmentation by detecting bounding boxes then followed by pixel level segmentation within each bounding box, for example, the well-known Mask R-CNN~\citep{he2017mask}. It first detects objects and then uses a mask branch and RoI-Align to segment instances within the proposed boxes. To better exploit the spatial information inside the box, PANet~\citep{wang2019panet} introduces bottom-up path augmentation, adaptive feature pooling and fuses mask predictions from fully-connected layers and convolutional layers. Such two-stage approaches achieve state-of-the-art performance. 
One stage instance segmentation methods are free of region proposals. In these methods, models output the pixel-wise auxiliary information, then a clustering algorithm groups information into object instances. Deep Watershed Transform~\citep{bai2017deep} predicts the energy map of the whole image and uses the watershed transform algorithm for grouping object instances. YOLACT~\citep{bolya2019yolact} generates prototype masks and the linear combination coefficients for each instance. Then linearly combines the prototype masks by corresponding coefficients to predict the instance-level mask.

\subsection{Contour-based Segmentation}
PolarMask~\citep{xie2020polarmask} uses polar representation to model the object boundary. The model is trained to simultaneously regress the object centroid as well as the length of 36 rays emitting uniformly with the same angle interval from the object centroid. Combined with the proposed Polar IoU Loss, several regression targets can be trained as a whole and thus improves the segmentation result. Instead of directly regressing the distance of the rays, ESE-Seg~\citep{xu2019explicit} proposes to further convert the polar coordinates of the object boundary to an explicit shape encoding using Chebyshev polynomial fitting, and turns the regression target to the coefficients of the Chebyshev polynomial. Although polar coordinates are inherently much suitable for modelling circular object boundary, both~\citep{xie2020polarmask,xu2019explicit} model the radial distance as a single-valued function of the polar angle, which makes it impossible to model the case of the ray  having multiple points intersect with the object boundary for a given angle. Curve-GCN~\citep{ling2019fast} is designed for assisting the manual annotation of class agnostic objects, it parametrizes object boundary with either polygons or splines, allowing annotation for both line-based and curved objects. They use graph convolutional network (GCN) to predict all the vertices of the polygon or control points of the spline along the object boundary in a iterative inference scheme. Deep Snake~\citep{peng2020deep} exploits the special topology of the object boundary as prior knowledge, and adopts circular convolution to predict the vertices along the object boundary. In order to obtain precise object boundary, both GCN and Deep Snake have to run inference multiple times. 
\section{Proposed Approach}
\subsection{Parametric Representation}
Parametric equations are commonly used to express the coordinates of points that make up a geometric object such as curve or surface. The general form of a parametric equation in two dimensions is shown in Eq.~(\ref{eq:parametric_equation}):
\begin{center}
\begin{equation}
\left\{
\begin{array}{ccc}
x & = & f(t) \\
y & = & g(t)
\end{array}
\right.
\label{eq:parametric_equation}
\end{equation}
\end{center}
where $t \in \mathbf{R}$ is the parameter and $f(t), g(t)$ are any explicit function of $t$. To recover the object shape, one can sample $t$ from the domain of the equation and obtain both $x$ and $y$ according to Eq.~(\ref{eq:parametric_equation}). Different from using a single-valued function to model the object shape~\citep{xie2020polarmask,xu2019explicit}, a parametric equation expresses the $x$ and $y$ coordinates of the object shape independently, which allows it to mimic a multi-valued function. This property provides much flexibilities compared to the single-valued function for shape representation.

\begin{figure*}[t!]
	\subfloat[\small{$t=0.25$}]{\includegraphics[width=0.33\textwidth]{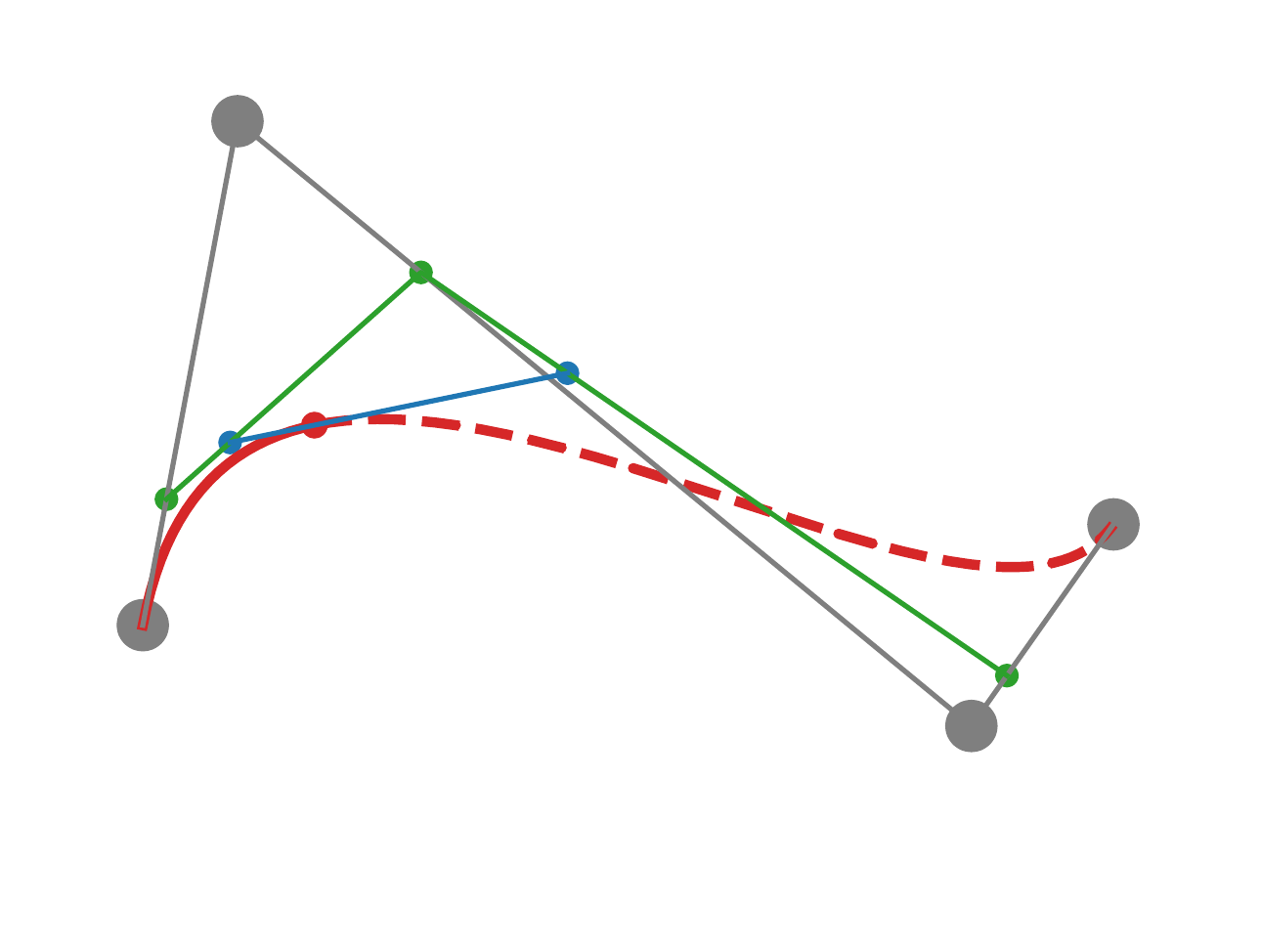}}
	\subfloat[\small{$t=0.50$}]{\includegraphics[width=0.33\textwidth]{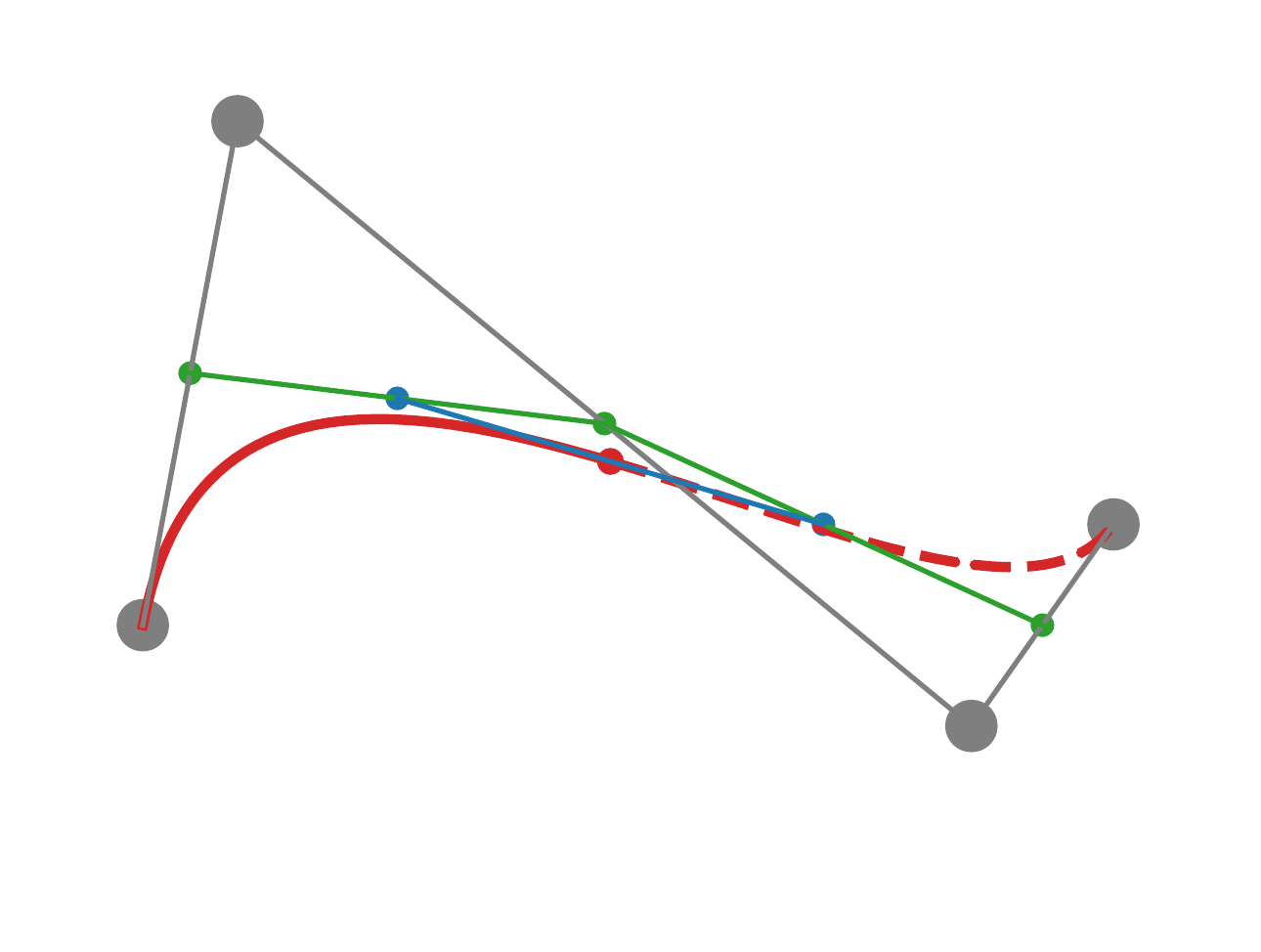}}
	\subfloat[\small{$t=0.75$}]{\includegraphics[width=0.33\textwidth]{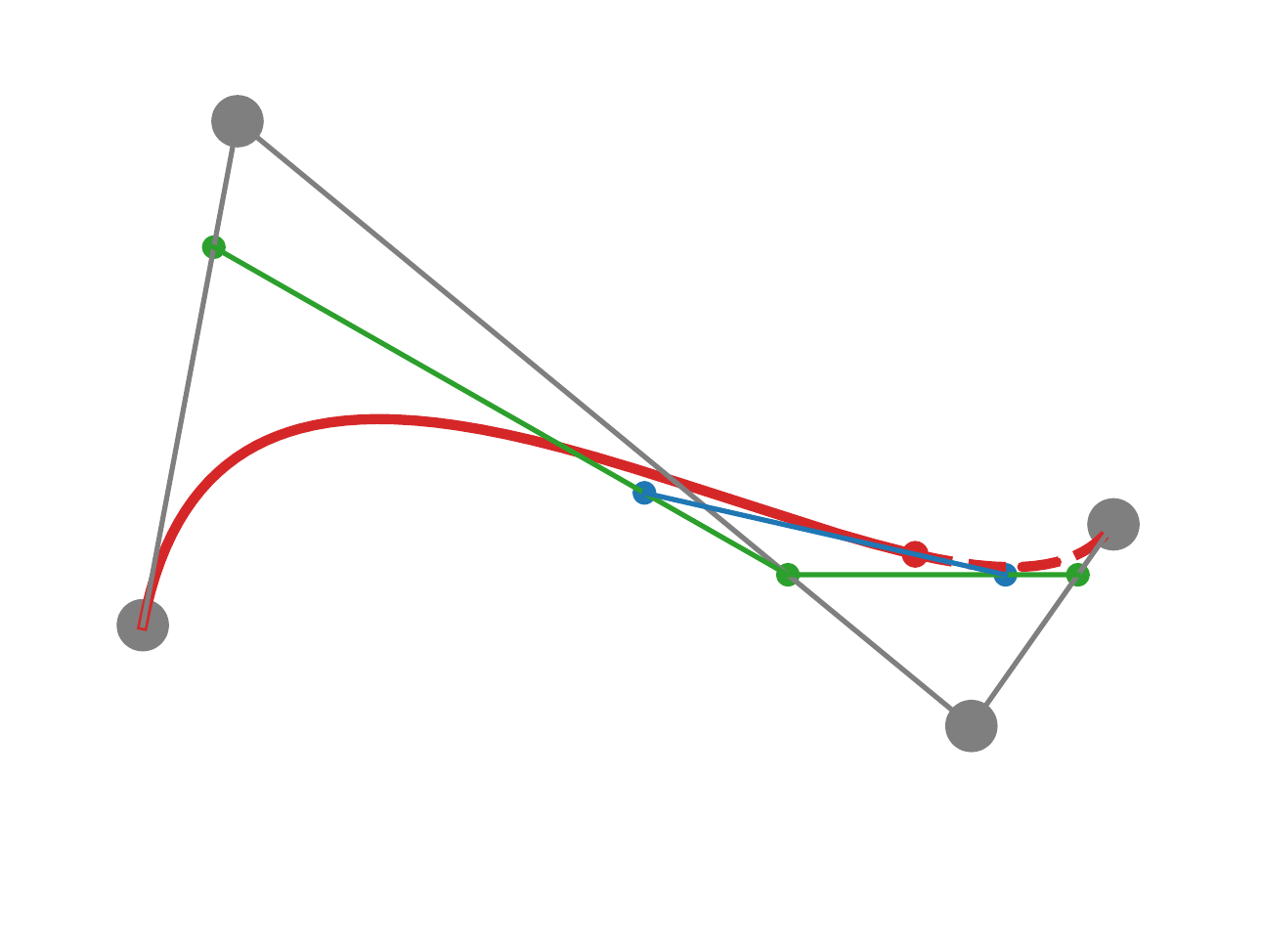}}
\caption{The process of constructing a cubic bezier curve. The trajectory of the red dot forms the curve as $t$ varies from 0 to 1.}
\label{fig:cubic_bezier_curve_animation}
\end{figure*}

\begin{figure}[t!]
\centering
\includegraphics[width=0.49\textwidth]{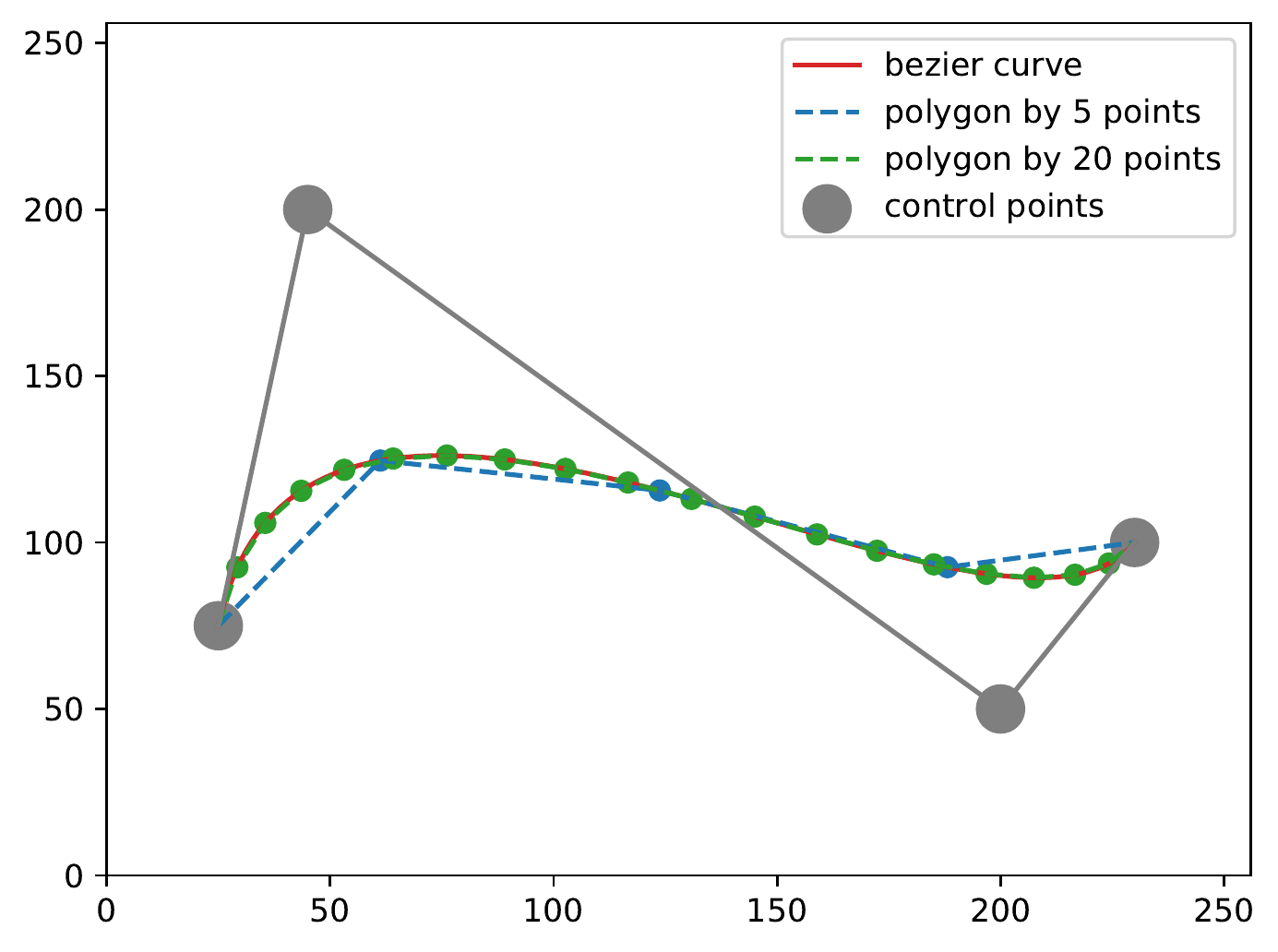}
\caption{Bezier curve vs polygon. The polygon representation requires more vertices in order to precisely describe the shape.}
\label{fig:bezier_curve_vs_polygon}
\end{figure}

Bezier curve is a set of parametric curve that have been widely used in many fields as an efficient design tool. The explicit definition of a bezier curve can be expressed as Eq.~(\ref{eq:def_bezier_curve}):
\begin{center}
\begin{equation}
\mathbf{B}(t) = \sum_{i=0}^{n} \binom{n}{i}(1 - t)^{n-i}t^i\mathbf{P}_i, 0 \le t \le 1
\label{eq:def_bezier_curve}
\end{equation}
\end{center}
where $\binom{n}{i}$ is the binomial coefficients, $\mathbf{P}_i = (x_i, y_i)$ is the $i$-$th$ control point of the bezier curve and $n$ is the degree of the bezier curve. One can construct a bezier curve by following these steps: 
\begin{enumerate}
\item Create the control polygon of the bezier curve by connecting the consecutive control points.
\item Insert intermediate points to each line segment, with the ratio $t:(1-t)$.
\item Treat the intermediate points as the new control points, repeat step 1), 2) until you end up with a single point.
\item As $t$ varies from 0 to 1, the trajectory of that single point forms the curve.
\end{enumerate} Fig.~\ref{fig:cubic_bezier_curve_animation} shows this process for constructing a cubic bezier curve. The use of bezier curve can reduce the number of parameters for shape encoding. As shown in Fig.~\ref{fig:bezier_curve_vs_polygon}, although the curve is determined by only four control points, it can guarantee the shape quality since the precision of the curve representation is fully depends on how densely you sample $\mathbf{t}$s from $[0, 1]$, whereas the polygon representation requires more vertices to achieve the similar precision. 

\begin{table}
\centering
\caption{Degree of bezier curve. Higher degree of bezier curve brings higher accuracy, while overmuch degree degrades the performance.}
\begin{tabular}{c|c|c|c|c}
\hline 
\textbf{Degree of bezier curve~($n$)} & 3 & 5 & 7 & 9 \\ 
\hline 
\textbf{MIOU} & 0.753 & 0.755 & 0.749 & 0.736 \\
\hline 
\end{tabular} 
\label{tab:num_points_performance}
\end{table}

\begin{figure*}[t!]
	\subfloat[]{\includegraphics[width=0.49\textwidth]{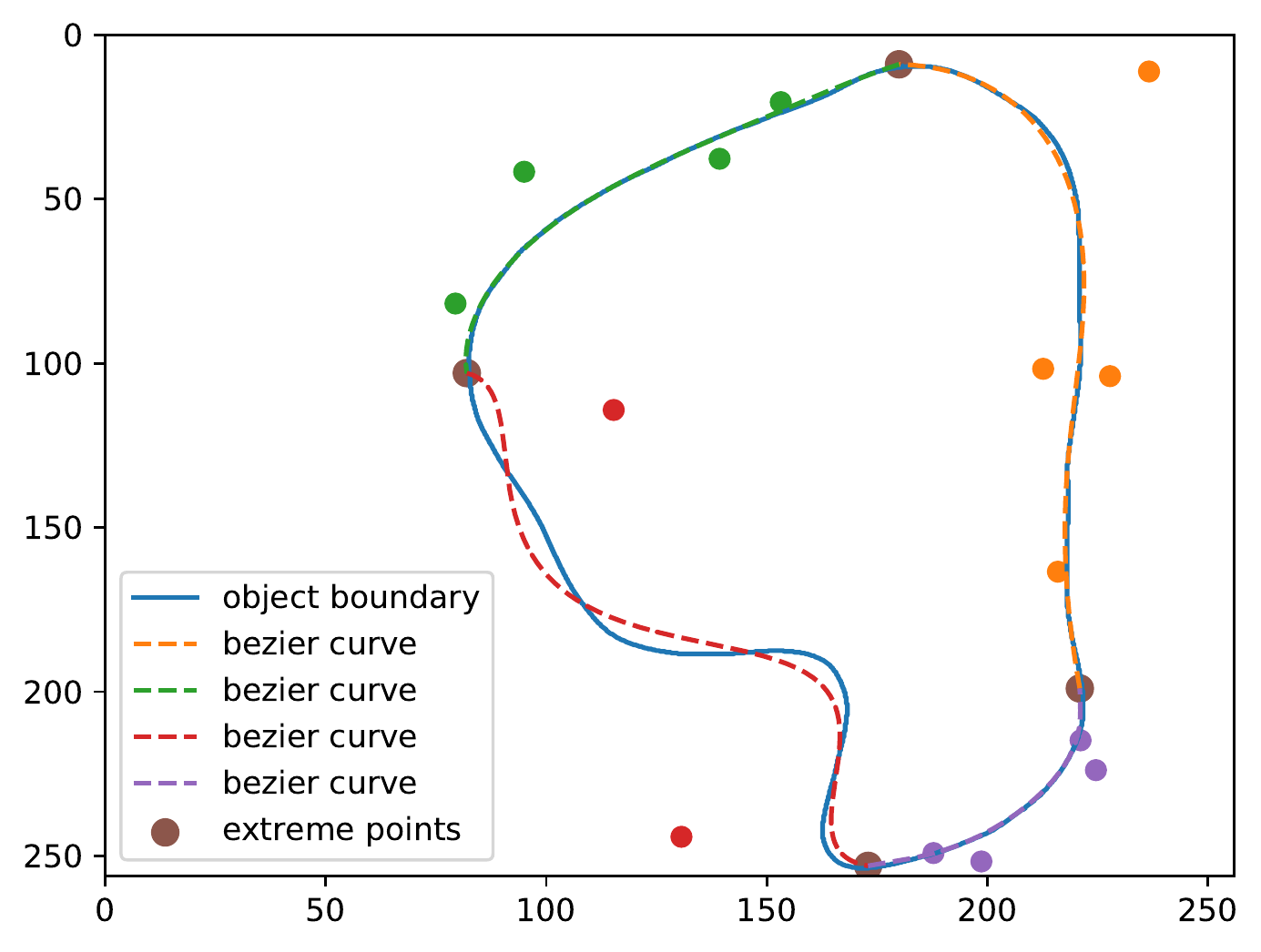}}
	\subfloat[]{\includegraphics[width=0.49\textwidth]{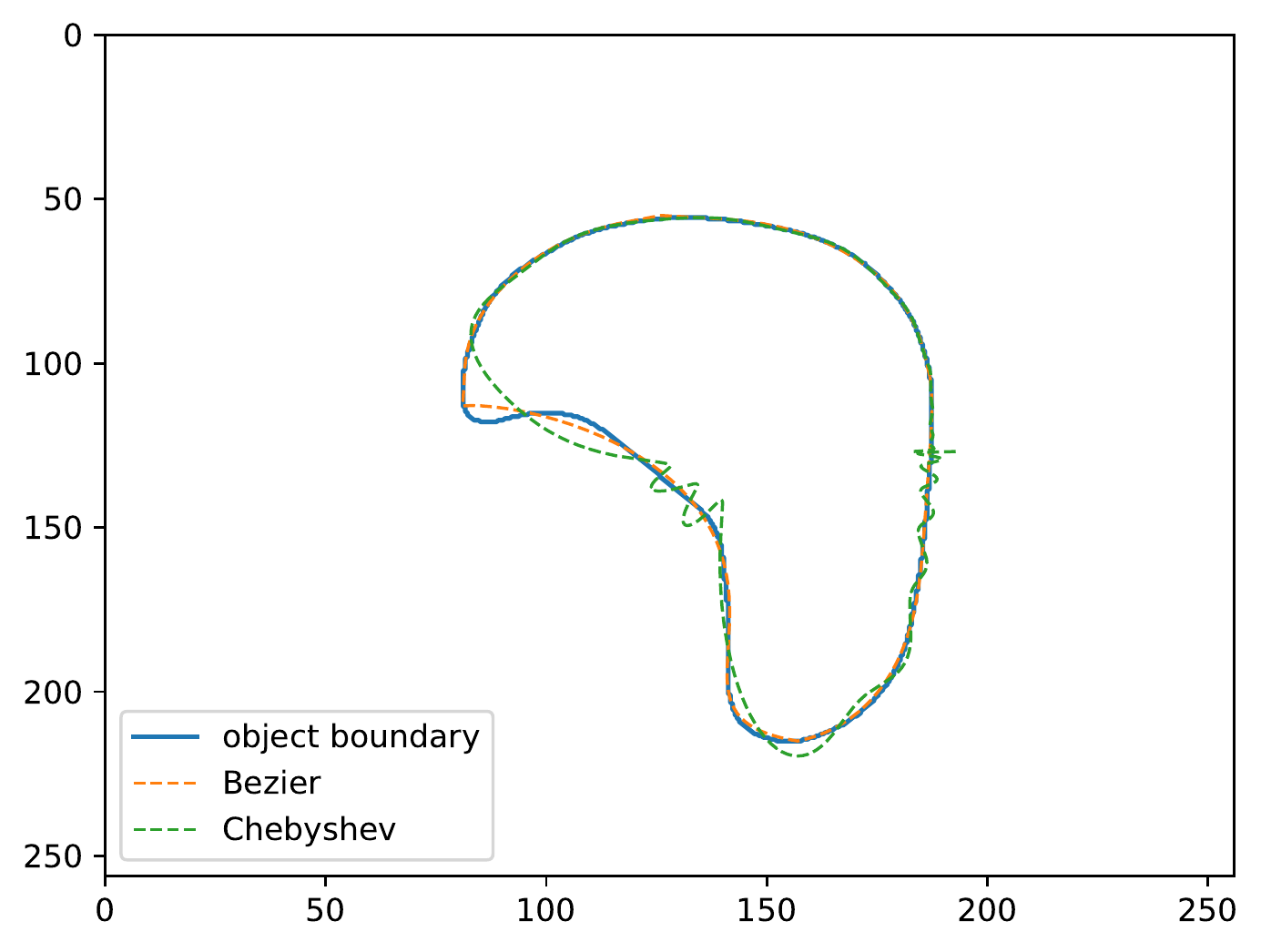}}
\caption{(a) Our piecewise bezier curves representation. (b) Our bezier curve representation can avoid the oscillation at the end of the object boundary compared to Chebyshev polynomial shape encoding.}
\label{fig:bezier_curves_representation}
\end{figure*}

Note that one can model the object shape using bezier curve in polar coordinates as well. In order to make the manually post refinement of the segmentation result much easier, we instead model the object shape in Cartesian coordinates. We conducted an experiment to study the accuracy of curve regression with respect to the degree of bezier curve and according to Table.~\ref{tab:num_points_performance}, we choose $n = 5$ for a trade-off between usability and segmentation accuracy. To retain higher accuracy after converting the object boundary to bezier curve representation, we propose to parametrize the object shape by piecewise bezier curves. Specifically, we first split the whole object boundary by its four extreme points, i.e., the top, the leftmost, the bottom and rightmost points of the object. Then for each part of the object boundary, we parametrize it with a bezier curve. Fig.~\ref{fig:bezier_curves_representation}~(a) gives an example of our bezier curves representation. We also compare our bezier curve representation with the Chebyshev polynomial shape encoding proposed in~\citep{xu2019explicit}. In Fig.~\ref{fig:bezier_curves_representation}~(b) we found that our bezier curve representation can avoid the oscillation at the end of the object boundary compared to the Chebyshev polynomial shape encoding.

\subsection{Model Architecture}

\begin{figure*}[t!]
\centering
\includegraphics[width=\textwidth]{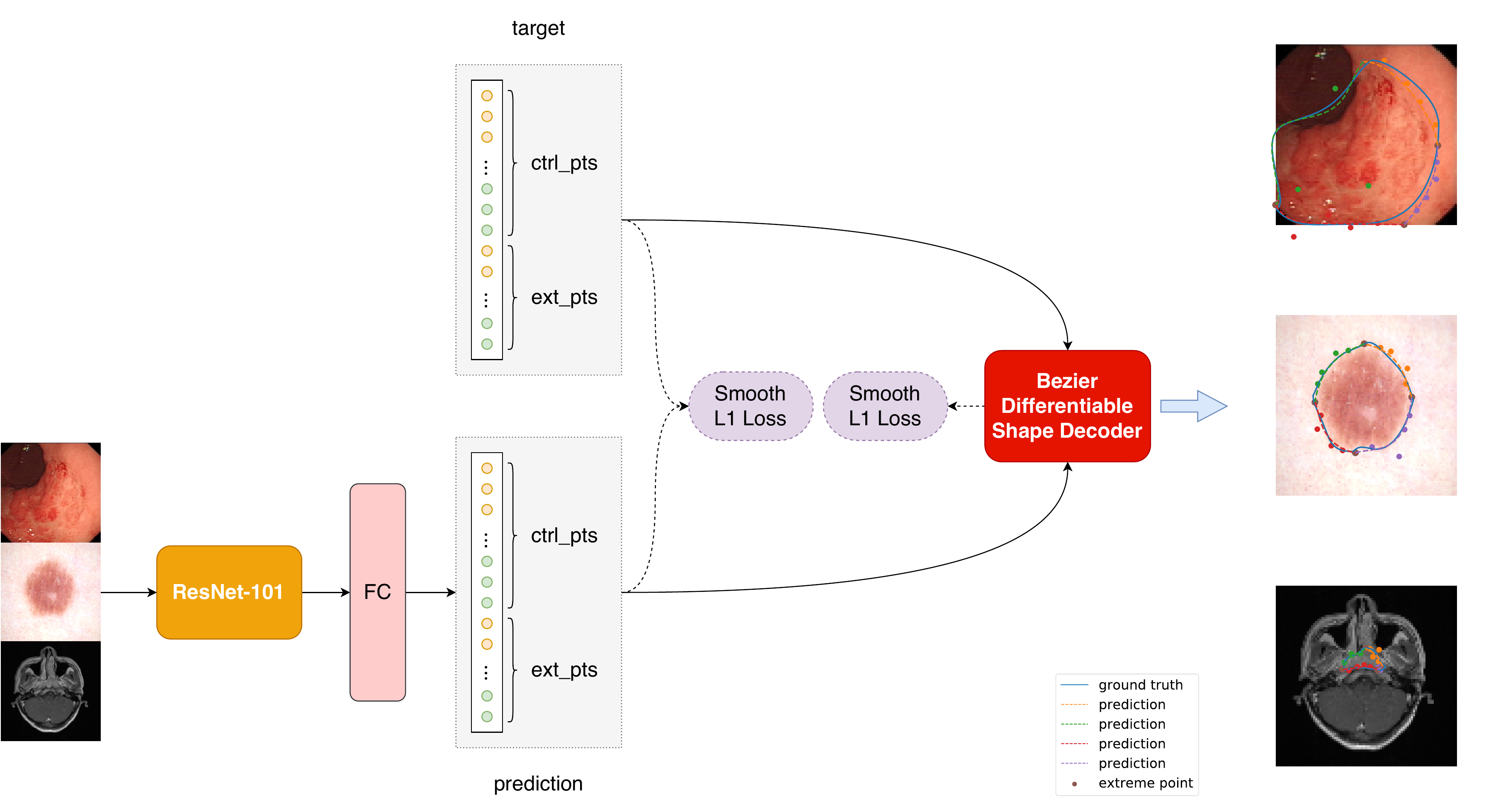}
\caption{Model architecture. \emph{ext\_pts} and \emph{ctrl\_pts} are the coordinates for $n_e$ extreme points and $n_c$ control points. Image features extracted from the ResNet-101 are passed to the fully connected~(FC) layer to obtain $(n_e+n_c)*2$ outputs. Thus the predicted coordinates of \emph{ext\_pts} and \emph{ctrl\_pts} regress to the target \emph{ext\_pts} and \emph{ctrl\_pts} supervised by smooth L1 loss denoted $\mathcal{L}_{ce}$. The bezier differentiable shape decoder module reconstructs $N$ predicted boundary points and $N$ ground truth boundary points. $\mathcal{L}_{matching}$ is their smooth L1 distance (see Sec.~\ref{sec:BDSD} and Sec.~\ref{sec:Loss}).}
\label{fig:model_arch}
\end{figure*}
As shown in Fig.~\ref{fig:model_arch}, we adopt ResNet-101~\citep{he2016deep} as the backbone of our model. We simply remove the softmax activation layer, and change the number of output nodes in the last fully connected layer to match the regression targets. For example, if $n_e$ extreme points and $n_c$ control points are needed, the output of the last fully connected layer should be $(n_e+n_c) * 2$ nodes, without any activation layer. Note that the idea in this paper can be easily applied to the mainstream object detection frameworks, which provides an alternative for instance segmentation.
\subsection{Ground Truth Label Generation}
\begin{figure}[t!]
\centering
\includegraphics[width=0.49\textwidth]{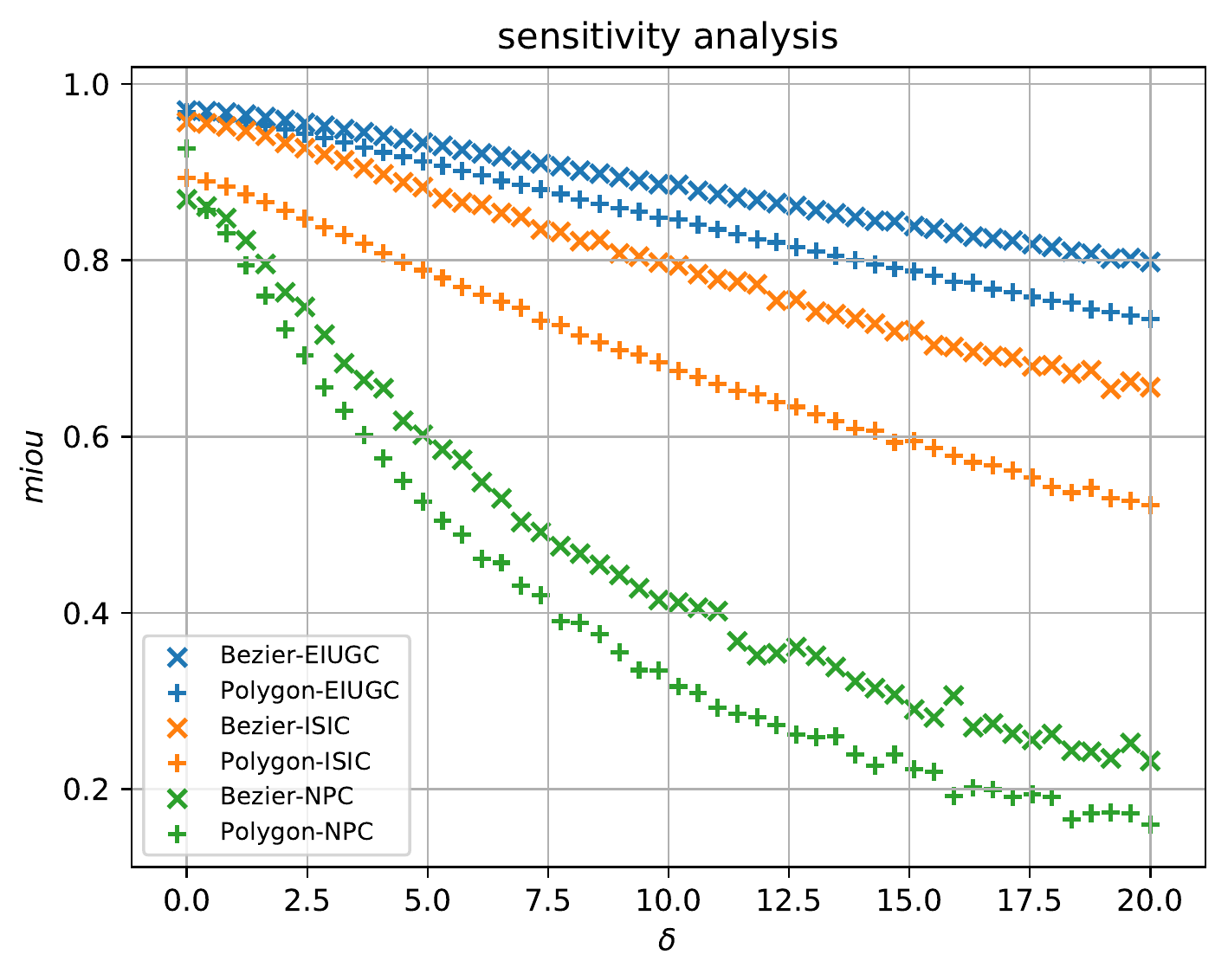}
\caption{Sensitivity analysis. The result of the analysis proved that the bezier curve representation is more robust than polygon representation under different noise levels.}
\label{fig:sensitivity_analysis}
\end{figure}

Given a mask of an object, we first extract the object boundary from the mask. Then, we find four extreme points of the object and use them  to split the whole shape into four parts. If there exist multiple extreme points such as two or more top points, we use the top left corner point, same as the bottom left one, the bottom right one and the top right one for leftmost, bottom, rightmost extreme point respectively. For each part of the object boundary, we fit it with a fifth-order bezier curve. The first and last control points of the bezier curve are set to be the begin point and end point of that piece of shape, which are two consecutive extreme points as well. The four intermediate control points of the bezier curve remain unknown. We convert Eq.~(\ref{eq:def_bezier_curve}) to the matrix form and place the coordinates of the object boundary part into the equation, as shown in Eq.~(\ref{eq:bezier_equation}):

\begin{center}
\begin{equation}
%\begin{array}{cccc} \left[\begin{array}{ccccc} C_1^1, & C_1^2, & C_1^3, & \cdots, & C_1^n \\ C_2^1, & C_2^2, & C_2^3, & \cdots, & C_2^n \\ C_3^1, & C_3^2, & C_3^3, & \cdots, & C_3^n \\ \vdots & \vdots & \vdots & \ddots & \vdots \\ C_m^1, & C_m^2, & C_m^3, & \cdots, & C_m^n \end{array}\right] & \left[\begin{array}{cc} x_1, & y_1 \\ x_2, & y_2 \\ x_3, & y_3 \\ \vdots & \vdots \\ x_n, & y_n \end{array}\right] & = & \left[\begin{array}{cc} \hat{x_1}, & \hat{y_1} \\ \hat{x_2}, & \hat{y_2} \\ \hat{x_3}, & \hat{y_3} \\ \vdots & \vdots \\ \hat{x_m}, & \hat{y_m} \end{array}\right]\end{array}
\begin{bmatrix} C_1^1, & C_1^2, & C_1^3, & \cdots, & C_1^n \\ C_2^1, & C_2^2, & C_2^3, & \cdots, & C_2^n \\ C_3^1, & C_3^2, & C_3^3, & \cdots, & C_3^n \\ \vdots & \vdots & \vdots & \ddots & \vdots \\ C_m^1, & C_m^2, & C_m^3, & \cdots, & C_m^n \end{bmatrix} \begin{bmatrix} x_1, & y_1 \\ x_2, & y_2 \\ x_3, & y_3 \\ \vdots & \vdots \\ x_n, & y_n \end{bmatrix} = \begin{bmatrix} \hat{x_1}, & \hat{y_1} \\ \hat{x_2}, & \hat{y_2} \\ \hat{x_3}, & \hat{y_3} \\ \vdots & \vdots \\ \hat{x_m}, & \hat{y_m} \end{bmatrix}
%\begin{array}{cccc} \left[\begin{array}{ccccc} \binom{n}{0}(1-t_0)^n, & \binom{n}{1}(1-t_0)^{n-1}t_0, & \binom{n}{2}(1-t_0)^{n-2}t_0^2, & \cdots, &\binom{n}{n}t_0^n \\ \binom{n}{0}(1-t_1)^n, & \binom{n}{1}(1-t_1)^{n-1}t_1, & \binom{n}{2}(1-t_1)^{n-2}t_1^2, & \cdots, & \binom{n}{n}t_1^n \\ \binom{n}{0}(1-t_2)^n, & \binom{n}{1}(1-t_2)^{n-1}t_2, & \binom{n}{2}(1-t_2)^{n-2}t_2^2, & \cdots, & \binom{n}{n}t_2^n \\ \vdots & \vdots & \vdots & \ddots & \vdots \\ \binom{n}{0}(1-t_m)^n, & \binom{n}{1}(1-t_m)^{n-1}t_m, & \binom{n}{2}(1-t_m)^{n-2}t_m^2, & \cdots, & \binom{n}{n}t_m^n \end{array}\right] &\left[\begin{array}{c}x_0 \\ x_1 \\ x_2 \\ \vdots \\ x_n \end{array}\right] &= &\left[\begin{array}{c} \hat{x_0} \\ \hat{x_1} \\ \hat{x_2} \\ \vdots \\ \hat{x_m} \end{array}\right]\end{array}
\label{eq:bezier_equation}
\end{equation}
\end{center}
in which $m$ represents the total number of points of the object boundary part, $n$ represents the number of control points of the bezier curve, $(x_i, y_i)$ is the $i$-$th$ control point, $(\hat{x_i}, \hat{y_i})$ is the $i$-$th$ points of the object boundary part and $C_i^j = \binom{n}{j}(1-t_i)^{n-j}t_i^j, t_i \in [0, 1]$. Inspired by the De Casteljau's algorithm for constructing a bezier curve, we assume that $t_i = (i - 1) / (m - 1)$ for the $i$-$th$ point of the object boundary part. Since the first and the last row of Eq.~(\ref{eq:bezier_equation}) are identity expression representing the extreme points, the first and last column are constant values, we remove the first and the last row from Eq.~(\ref{eq:bezier_equation}) and subtract the constant values from both side of Eq.~(\ref{eq:bezier_equation}). To simplified the notation, let:
\begin{center}
\begin{gather*}
%\begin{array}{c} \mathbf{A} = \left[\begin{array}{ccccc} C_2^2, & C_2^3, & C_2^4, & \cdots, & C_2^{n-1} \\ C_3^2, & C_3^3, & C_3^4, & \cdots, & C_3^{n-1} \\ C_4^2, & C_4^3, & C_4^4, & \cdots, & C_4^{n-1} \\ \vdots & \vdots & \vdots & \ddots & \vdots \\ C_{m-1}^2, & C_{m-1}^3, & C_{m-1}^4, & \cdots, & C_{m-1}^{n-1} \end{array}\right], \mathbf{c} = \left[\begin{array}{cc} x_2, & y_2 \\ x_3, & y_3 \\ x_4, & y_4 \\  \vdots & \vdots \\ x_{n-1}, & y_{n-1} \end{array}\right], \\ \\ \mathbf{b} = \left[\begin{array}{cc} \hat{x_2}-C_2^1 x_1-C_2^n x_n, & \hat{y_2}-C_2^1 y_1-C_2^n y_n \\ \hat{x_3}-C_3^1 x_1-C_3^n x_n, & \hat{y_3}-C_3^1 y_1-C_3^n y_n \\ \hat{x_4}-C_4^1 x_1-C_4^n x_n, & \hat{y_4}-C_4^1 y_1-C_4^n y_n \\ \vdots & \vdots \\ \hat{x_{m-1}}-C_{m-1}^1 x_1-C_{m-1}^n x_n, & \hat{y_{m-1}}-C_{m-1}^1 y_1-C_{m-1}^n y_n \end{array}\right] \end{array}
\tiny{\mathbf{A} = \begin{bmatrix} C_2^2, & C_2^3, & C_2^4, & \cdots, & C_2^{n-1} \\ C_3^2, & C_3^3, & C_3^4, & \cdots, & C_3^{n-1} \\ C_4^2, & C_4^3, & C_4^4, & \cdots, & C_4^{n-1} \\ \vdots & \vdots & \vdots & \ddots & \vdots \\ C_{m-1}^2, & C_{m-1}^3, & C_{m-1}^4, & \cdots, & C_{m-1}^{n-1} \end{bmatrix}}, \\ \tiny{\mathbf{b} = \begin{bmatrix} \hat{x_2}-x_1 C_2^1-x_n C_2^n, & \hat{y_2}-y_1 C_2^1-y_n C_2^n \\ \hat{x_3}-x_1 C_3^1-x_n C_3^n, & \hat{y_3}-y_1 C_3^1-y_n C_3^n \\ \hat{x_4}-x_1 C_4^1-x_n C_4^n, & \hat{y_4}-y_1 C_4^1-y_n C_4^n \\ \vdots & \vdots \\ \hat{x_{m-1}}-x_1 C_{m-1}^1-x_n C_{m-1}^n, & \hat{y_{m-1}}-y_1 C_{m-1}^1-y_n C_{m-1}^n \end{bmatrix}}, \\ \tiny{\mathbf{c} = \begin{bmatrix} x_2, & y_2 \\ x_3, & y_3 \\ x_4, & y_4 \\  \vdots & \vdots \\ x_{n-1}, & y_{n-1} \end{bmatrix}}
\end{gather*}
\end{center}
thus $\mathbf{A}\mathbf{c} = \mathbf{b}$. By computing the pseudo inverse of $\mathbf{A}$ and multiply it to both side of the equation, we can obtain the coordinates of the intermediate control points $\mathbf{c} = \mathbf{A}^{-1}\mathbf{b}$. Finally, we concatenate the coordinates of four extreme points and 16 intermediate control points as our regression target ($n_e = 4, n_c = 16$), resulting in a 40 dimensional vector for each input image. Similar to ESE-Seg, we also conduct sensitivity analysis for our bezier curve representation. Specifically, we randomly sample some noises from $N(0, \delta)$ and add them to the control points as well as the extreme points to imitate the uncertainty behaviour of the convolutional neural network. We compare our bezier curve representation with the polygon shape representation. To maintain the same complexity for both shape representation, we evenly select 20 points along the object boundary for the polygon representation. Fig.~\ref{fig:sensitivity_analysis} shows the robustness of our bezier curves representation. As shown in Fig.~\ref{fig:sensitivity_analysis}, bezier curves representation always achieves higher Mean Intersection-Over-Union (MIOU) than polygon representation with the same number of points. We can conclude that bezier curves representation is superior than polygon representation because it allows us to describe a curve more precisely. The bezier curves representation is robust for the endoscopy images of upper gastrointestinal cancers (\textit{EIUGC}) dataset and the international skin imaging collaboration skin lesion challenge (\textit{ISIC}) dataset, remaining above 0.6 miou even the noise of $\delta=20$ was injected. For the same perturbation, the miou of the nasopharyngeal carcinoma magnetic resonance imaging (\textit{NPCMRI}) dataset dropped rapidly mainly due to its smaller average size of ROI comparing to the other two datasets. 
\subsection{Bezier Differentiable Shape Decoder}
\label{sec:BDSD}
Similar to FourierNet~\citep{riaz2021fouriernet}, We devise a novel differentiable shape decoder named \emph{Bezier Differentiable Shape Decoder}, abbreviated as \emph{BDSD}. FourierNet used Inverse Fast Fourier Transformation as shape decoder to convert the coefficients of the Fourier series into contour points whereas we apply the parametric equation of bezier curve to map the control points to contour points. Following Eq.~(\ref{eq:def_bezier_curve}), we randomly sample $N$ $\mathbf{t}$s from $U[0, 1]$ and use the same $\mathbf{t}$s to reconstruct $N$ ground truth boundary points and the corresponding predicted boundary points. In all experiments, we set $N = 72$. Please note that the \emph{BDSD} module is fully differentiable and allows the gradient to back propagate through the network and thus can provide extra supervision during the training phase.
\subsection{Loss}
\label{sec:Loss}
The overall loss function for training our model contains two individual loss terms. The first one $\mathcal{L}_{ce}$ is the Smooth L1 loss for control points and extreme points regression, and the second one $\mathcal{L}_{matching}$ is another Smooth L1 loss for point matching learning, which measures the differences between the outputs of the \emph{BDSD} module. Eq.~(\ref{eq:loss}) shows the overall loss function:
\begin{equation}
\mathcal{L} = \lambda_{ce} \mathcal{L}_{ce} + \lambda_{matching} \mathcal{L}_{matching}
\label{eq:loss}
\end{equation}
where $\lambda_{ce}$ and $\lambda_{matching}$ are balancing hyper parameters, we set both of them equal to 1 in all experiments.

\section{Experiments}

\subsection{Datasets and Evaluation Metric}
Endoscopy images of upper gastrointestinal cancers (\textit{EIUGC}) is an upper gastrointestinal endoscopy image dataset, which aims to detect upper gastrointestinal cancers. \textit{EIUGC} collected 38,453 endoscopy images. The border of each cancer lesion was marked by highly experienced endoscopists from SYSUCC. We separate the dataset into a training set, a validation set and a test set with 30762, 3845 and 3846 images respectively.

The nasopharyngeal carcinoma magnetic resonance imaging (\textit{NPCMRI}) dataset enrolled 375 patients at SYSUCC from January 2012 to December 2014. The NCMRI contains multi-parametric MRI sequences of nasopharyngeal carcinoma patients, including T1w, T2w, T1c. Doctors delineated the invasive ROIs of NPC in T1c by referencing T1w, T2w sequences. As a result, we used the T1c sequence as source images. Totally 2,337 images were collected.

The international skin imaging collaboration (\textit{ISIC}) skin lesion challenge datasets of 2018 was released in ISIC Challenge~\citep{isic_training}, which aimed at skin lesion analysis towards melanoma detection. One of its three tasks is lesion segmentation. The dataset includes 2,576 skin images and their corresponding masks, of which 2060 in training set, 258 in validation set and 258 in test set. All ground truth data were reviewed and curated by practicing dermatologists with expertise in dermoscopy. The image may be associated with multiple mask annotations. In that case, we randomly select one of the expert annotation and fall back to use novice annotation if no expert annotation provided. The above three datasets are summarized in Table.~\ref{tab:dataset_summaries}. 

We use Mean Intersection-Over-Union~(miou), Hausdorff Distance~\citep{huttenlocher1993comparing}, Matthews Correlation Coefficient~\citep{matthews1975comparison}~(mcc), Area Under Curve~(auc), False Positive Rate~(fp) and False negative Rate~(fn) to evaluate the proposed method, a contour-based model, PolarMask ResNet101 and a pixel-based model, DeepLab v3+ ResNet101 on \textit{EIUGC}, \textit{NPCMRI} and \textit{ISIC}.

\begin{table}
\centering
\caption{Dataset overview. The \emph{MIOU} and \emph{SIOU} in the table represents the mean intersection-over-union and the standard deviation of the intersection-over-union between the bezier curve representation and the ground truth mask respectively.}
\begin{tabular}{c|c|c|c|c|c}
\hline 
\textbf{dataset} & \textbf{trainset} & \textbf{valset} & \textbf{testset} & \textbf{MIOU} & \textbf{SIOU} \\ 
\hline 
\textit{EIUGC} & 30762 & 3845 & 3846 & 0.970 & 0.019 \\ 
\textit{NPCMRI} & 1869 & 234 & 234 & 0.869 & 0.056 \\ 
\textit{ISIC} & 2060 & 258 & 258 & 0.957 & 0.021 \\ 
\hline 
\end{tabular} 
\label{tab:dataset_summaries}
\end{table}

\begin{figure}[h!]
\centering
\includegraphics[width=0.49\textwidth]{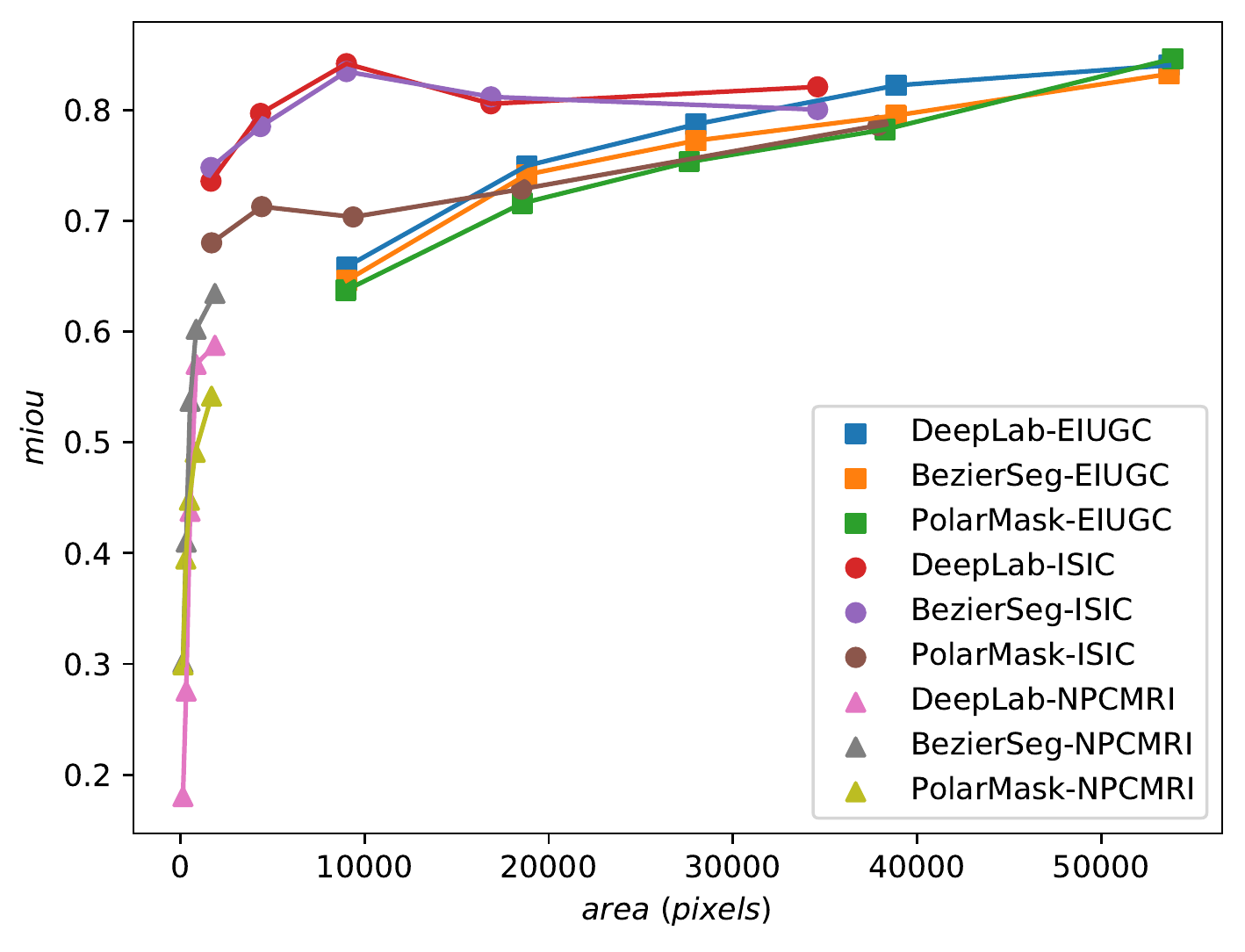}
\caption{Performance comparison between \textbf{BezierSeg}, \textbf{PolarMask} and \textbf{DeepLab v3+} on different area range}
\label{fig:iou-area}
\end{figure}

\subsection{Results}
\begin{figure*}
\centering
\includegraphics[width=\textwidth]{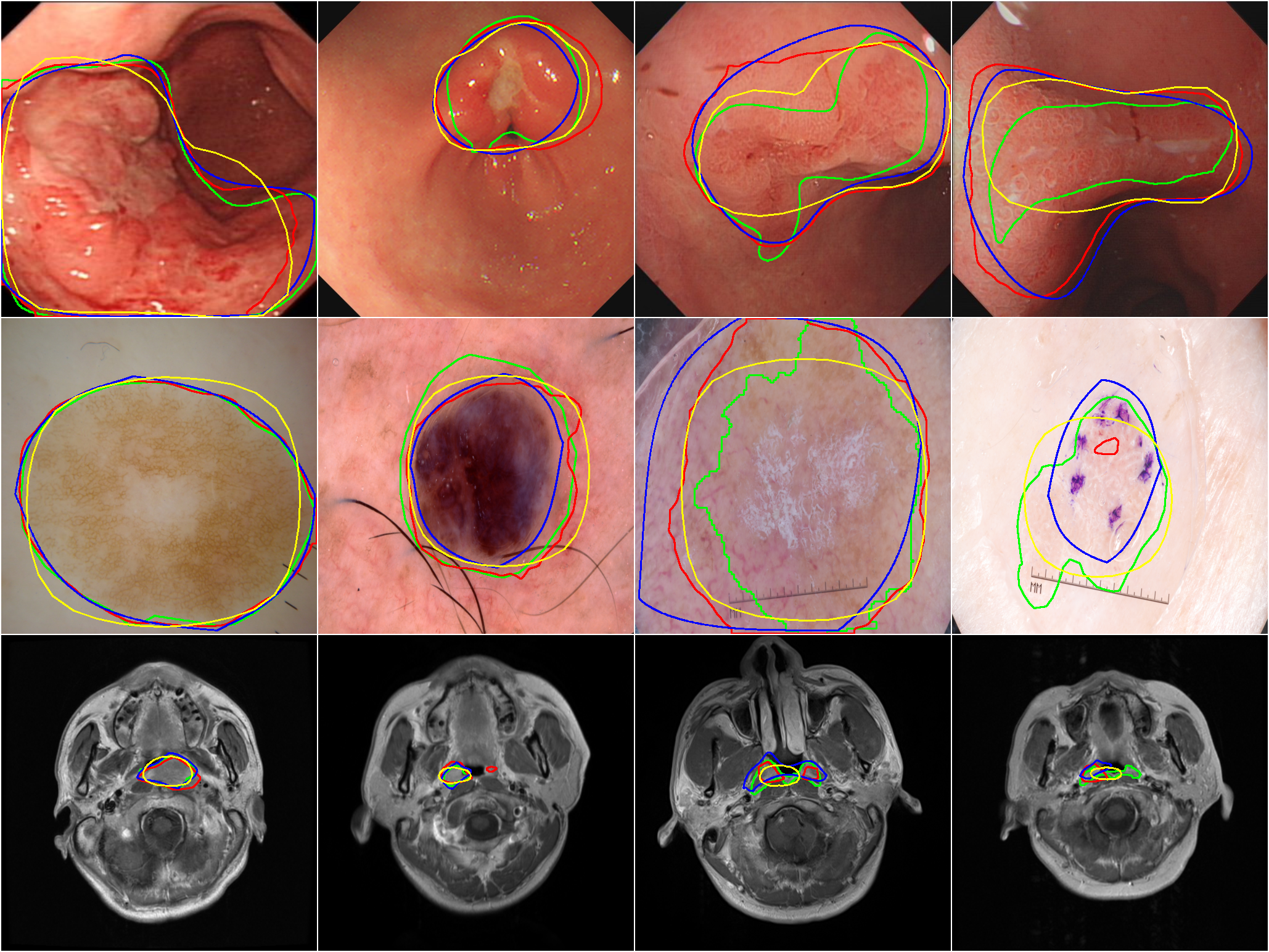}
\caption{Qualitative results comparison between \textbf{BezierSeg}, \textbf{DeepLab v3+ ResNet101} and \textbf{PolarMask} on three datasets. The first row, second row and the last row show the results on the \textit{EIUGC} dataset, the \textit{ISIC} dataset and the \textit{NPCMRI} dataset respectively. Green: Ground-truth label; Red: \textbf{DeepLab v3+ ResNet101}; Blue: \textbf{BezierSeg}; Yellow: \textbf{PolarMask}.}
\label{fig:qualitative-results}
\end{figure*}

As shown in Table.~\ref{tab:exp_res}, the segmentation performance of BezierSeg 50 is slightly worse than BezierSeg 101, we therefore use ResNet-101 as the backbone of our model by default. Without any bells and whistles, BezierSeg can achieve competitive result with pixel-based methods. On both \textit{EIUGC} and \textit{ISIC} datasets, the differences of miou between these two methods are less than 0.02. The average ROI sizes in \textit{EIUGC}, \textit{NPCMRI} and \textit{ISIC} are 29337 pixels, 14152 pixels and 745 pixels respectively. BezierSeg outperforms DeepLab v3+ ResNet101 on \textit{NPCMRI} by 21\% and behind DeepLab v3+ on \textit{EIGUC} and \textit{ISIC} by only 1.8\% and 0.6\% respectively. In terms of hausdorff distance, BezierSeg outperforms DeepLab v3+ on both \textit{ISIC} and \textit{NPC}. PolarMask performs relatively poorly on all three datasets.  From Fig.~\ref{fig:iou-area} we can see that when the ROI size is greater than 20k, DeepLab v3+ ResNet101 performs slightly better; when the ROI size is between 4k and 20k, the proposed model and DeepLab v3+ ResNet101 perform similarly; when the ROI size is less than 4k , Both BezierSeg and PolarMask outperforms DeepLab v3+ ResNet101. The overall performance of BezierSeg is less sensitive to the size of ROI comparing to DeepLab v3+ ResNet101. As show in Fig.~\ref{fig:qualitative-results}, BezierSeg can always output a smooth contour, whereas DeepLab v3+ ResNet101 outputs rugged contour and often produces multiple separated contours. PolarMask gives polygonal outputs that roughly ensures the smoothness, however, it is difficult to handle the dumbbell shape and the model outputs pebble shape in most of that case.

\begin{table*}[t!]
\centering
\caption{Comparison of BezierSeg ResNet50~(BezierSeg 50), BezierSeg ResNet101~(BezierSeg 101), DeepLab v3+ ResNet101~(DeepLab v3+) and PolarMask ResNet101~(PolarMask) on three datasets. \textit{curve miou} is the mean intersection-over-union between the prediction and the ground truth bezier curves shape representation, \textit{mask miou} is the mean intersection-over-union between the prediction and the ground truth mask.}
\begin{tabular}{c|c|c|c|c|c|c|c|c|c}
\hline 
\rule[-1ex]{0pt}{2.5ex} dataset & model & BDSD & curve miou & mask miou & hausdorff & mcc & auc & fp & fn \\ 
\hline 
\rule[-1ex]{0pt}{2.5ex} \multirow{5}{*}{\textit{EIUGC}} & DeepLab v3+ & $-$ & $-$ & \textbf{0.772} & \textbf{14.330} & \textbf{0.781} & \textbf{0.891} & 0.104 & 0.115 \\ 
\cline{2-10} 
\rule[-1ex]{0pt}{2.5ex} ~ & PolarMask & $-$ & $-$ & 0.747 & 16.263 & 0.753 & 0.878 & 0.129 & 0.116 \\ 
\cline{2-10}
\rule[-1ex]{0pt}{2.5ex} ~ & BezierSeg 50 & $\circ$ & 0.750 & 0.745 & 15.473 & 0.752 & 0.876 & 0.118 & 0.129 \\ 
\cline{2-10}  
\rule[-1ex]{0pt}{2.5ex} ~ & BezierSeg 50 & \checkmark & 0.750 & 0.745 & 15.573 & 0.751 & 0.876 & 0.126 & 0.122 \\ 
\cline{2-10} 
\rule[-1ex]{0pt}{2.5ex} ~ & BezierSeg 101 & $\circ$ & 0.761 & 0.757 & 14.898 & 0.763 & 0.882 & 0.113 & 0.124 \\ 
\cline{2-10}  
\rule[-1ex]{0pt}{2.5ex} ~ & BezierSeg 101 & \checkmark & 0.762 & 0.758 & 14.818 & 0.763 & 0.882 & 0.119 & 0.117 \\ 
\hline 
\rule[-1ex]{0pt}{2.5ex} \multirow{5}{*}{\textit{ISIC}} & DeepLab v3+ & $-$ & $-$ & \textbf{0.801} & 8.139 & \textbf{0.877} & \textbf{0.939} & 0.028 & 0.093 \\ 
\cline{2-10}  
\rule[-1ex]{0pt}{2.5ex} ~ & PolarMask & $-$ & $-$ & 0.723 & 8.707 & 0.812 &0.903 & 0.039 & 0.156 \\ 
\cline{2-10}
\rule[-1ex]{0pt}{2.5ex} ~ & BezierSeg 50 & $\circ$ & 0.796 & 0.791 & \textbf{7.339} & 0.869 & 0.934 & 0.029 & 0.104 \\ 
\cline{2-10}  
\rule[-1ex]{0pt}{2.5ex} ~ & BezierSeg 50 & \checkmark & 0.799 & 0.793 & 7.349 & 0.872 & 0.936 & 0.029 & 0.099 \\ 
\cline{2-10}  
\rule[-1ex]{0pt}{2.5ex} ~ & BezierSeg 101 & $\circ$ & 0.797 & 0.792 & 7.414 & 0.866 & 0.934 & 0.031 & 0.102 \\ 
\cline{2-10}  
\rule[-1ex]{0pt}{2.5ex} ~ & BezierSeg 101 & \checkmark & 0.801 & 0.796 & 7.406 & 0.870 & 0.936 & 0.030 & 0.097 \\
\hline 
\rule[-1ex]{0pt}{2.5ex} \multirow{5}{*}{\textit{NPCMRI}} & DeepLab v3+ & $-$ & $-$ & 0.413 & 9.553 & 0.671 & 0.820 & 0.003 & 0.357 \\ 
\cline{2-10}  
\rule[-1ex]{0pt}{2.5ex} ~ & PolarMask & $-$ & $-$ & 0.437 & 5.348 & 0.650 & 0.799 & 0.003 & 0.398 \\ 
\cline{2-10}
\rule[-1ex]{0pt}{2.5ex} ~ & BezierSeg 50 & $\circ$ & 0.482 & 0.467 & 5.285 & 0.685 & 0.840 & 0.003 & 0.317 \\ 
\cline{2-10}  
\rule[-1ex]{0pt}{2.5ex} ~ & BezierSeg 50 & \checkmark & 0.504 & 0.488 & 4.909 & 0.707 & 0.855 & 0.003 & 0.287 \\
\cline{2-10}  
\rule[-1ex]{0pt}{2.5ex} ~ & BezierSeg 101 & $\circ$ & 0.500 & 0.478 & 5.258 & 0.696 & 0.856 & 0.004 & 0.284 \\ 
\cline{2-10}  
\rule[-1ex]{0pt}{2.5ex} ~ & BezierSeg 101 & \checkmark & 0.520 & \textbf{0.499} & \textbf{4.886} & \textbf{0.717} & \textbf{0.866} & 0.003 & 0.265 \\ 
\hline 
\end{tabular}  
\label{tab:exp_res}
\end{table*}

We also compare the FPS between BezierSeg, PolarMask and DeepLab v3+ on the same machine equipped with one Tesla V100 graphics card. As shown in Table.~\ref{tab:speed_comp}, BezierSeg, which does not require upsampling layers, has 42.6M parameters while DeepLab v3+ ResNet101 has 58.6M. With simpler pipeline, BezierSeg doubles the FPS to DeepLab v3+ ResNet101. Both BezierSeg and PolarMask are lightweight models with similar architecture, they show no significant difference when running without post-processing. However, BezierSeg consumes more time to reconstruct the smooth contour during the post-processing stage. The speed of BezierSeg makes it more suitable for real time medical segmentation. A smaller size of model has more flexibility to transfer to edge-computing device.

\begin{table}[h!]
\centering
\caption{Speed comparison between DeepLab v3+ ResNet101~(DeepLab v3+), BezierSeg ResNet101~(BezierSeg) and PolarMask ResNet101~(PolarMask), including both with/without post-processing to reconstruct the object boundary.}
\begin{tabular}{c|c|c}
\hline
\rule[-1ex]{0pt}{2.5ex} model & post-processing & FPS \\ 
\hline 
\rule[-1ex]{0pt}{2.5ex} \multirow{2}{*}{DeepLab v3+} & $\circ$ & 48.4 \\ 
\cline{2-3} 
\rule[-1ex]{0pt}{2.5ex} ~ & \checkmark & 45.6 \\ 
\hline 
\rule[-1ex]{0pt}{2.5ex} \multirow{2}{*}{\textbf{BezierSeg}} & $\circ$ & \textbf{103.8} \\ 
\cline{2-3}
\rule[-1ex]{0pt}{2.5ex} ~ & \checkmark & \textbf{97.8} \\ 
\hline 
\rule[-1ex]{0pt}{2.5ex} \multirow{2}{*}{\textbf{PolarMask}} & $\circ$ & \textbf{103.3} \\ 
\cline{2-3}
\rule[-1ex]{0pt}{2.5ex} ~ & \checkmark & \textbf{101.9} \\ 
\hline 
\end{tabular} 
\label{tab:speed_comp}
\end{table}

\subsection{Implementation Details}
For training DeepLab v3+ ResNet101, we use binary cross entropy loss function and initialize the model with the weights pre-trained on COCO~\citep{lin2014microsoft} train2017, and for training BezierSeg, we initialize the backbone of our model with the weights pre-trained on ImageNet~\citep{deng2009imagenet}. As for the configuration of PolarMask, we use ResNet101 as backbone and use 36 rays for shape representation. During training, for the \textit{EIUGC} and the \textit{ISIC} datasets, we use data augmentation techniques such as random horizontal flipping or vertical flipping with a probabilities of 0.5, image rotation with a random angle ranging from $[-20, 20]$, image scaling with a random scale factor ranging from $[0.8, 1.25]$ followed by random fixed size image cropping. For the \textit{NPCMRI} dataset, we apply the same techniques above except that we replace the random image scaling and cropping with the fixed image resizing augmentation, and change the range of the rotation angle from $[-20, 20]$ to $[-10, 10]$. During the inference phase, we simply resize the input images to the size of $(256, 256)$. For both training and inference phases, we normalize the input images by dividing their pixel value by $255$. We dynamically generate both the extreme points and the control points after performing data augmentation. For the \textit{ISIC} dataset, we apply morphological transformations to the mask to remove the spikes along the object boundary before fitting the bezier curves. The initial learning rate is set to 1e-3 for all experiments. We also adopt adaptive learning rate scheduling strategy during training. Specifically, we reduce the learning rate by half when the loss of the validation set has stopped decreasing for 15 epochs and the monitoring threshold is set to 1e-2 and 0.5 for DeepLab v3+ ResNet101 and BezierSeg respectively. We train the models for 100 epochs in all experiments and evaluate the performance on the validation set at the end of each epoch. We pick up the model with highest miou on the validation set to evaluate the performance on the hold-out test set. The final results are obtained by running the experiments three times with different random seed and taking the average.
\section{Conclusions}
We propose the \emph{BezierSeg} model, which can directly output bezier curves without post-processing. Its simple pipeline provides the capability of real-time inference. In the experiments, it can be found that BezierSeg shows competitive accuracy in multiple medical datasets. However, our model fails to handle disconnected areas such as dilated shape contour due to the limitation of the 2D bezier curve representation. One possible solution for this case is to incorporate parametric surface representation such as level set. We will explore the potential application of bezier surface for 3D object segmentation in our future work, which is also a promising and practical segmentation solution for three-dimensional medical data such as computed tomography~(CT) and magnetic resonance~(MR) images.

\section*{Acknowledgment}

This work was supported by the National Natural Science Foundation of China under Grant No. 81702873, the Fundamental Research Funds for the Central Universities under Grant No. 19tkpy201. This study was approved by the Research Ethics Committee of Sun Yat-sen University Cancer Center (SYSUCC), and written informed consent was obtained from all patients before treatment. The key raw data have been uploaded onto the Research Data Deposit public platform (RDD), with the approval of RDD number RDDB2019000564 and RDDA2019001214 for the \textit{NPCMRI} dataset and the \textit{EIUGC} dataset respectively.

% Use the style of numbering in square brackets.
% If nothing is used, default style will be taken.
%\begin{enumerate}[a)]
%\item 
%\item 
%\item 
%\end{enumerate}  

%\begin{itemize}
%\item 
%\item 
%\item 
%\end{itemize}  

%\begin{description}
%\item[]
%\item[] 
%\item[] 
%\end{description}  

% Figure
%\begin{figure}[<options>]
%	\centering
%		\includegraphics[<options>]{}
%	  \caption{}\label{fig1}
%\end{figure}

%\begin{table}[<options>]
%\caption{}\label{tbl1}
%\begin{tabular*}{\tblwidth}{@{}LL@{}}
%\toprule
%  &  \\ % Table header row
%\midrule
% & \\
% & \\
% & \\
% & \\
%\bottomrule
%\end{tabular*}
%\end{table}

% Uncomment and use as the case may be
%\begin{theorem} 
%\end{theorem}

% Uncomment and use as the case may be
%\begin{lemma} 
%\end{lemma}


%\section{}\label{}


\bibliographystyle{cas-model2-names}

\bibliography{bibliography}



\end{document}